\documentclass[letterpaper, 10 pt, conference]{ieeeconf}  

\IEEEoverridecommandlockouts                              

\usepackage{cite}
\usepackage{amsmath,amssymb,amsfonts}
\usepackage{algorithm}
\usepackage{algpseudocode}
\usepackage{multirow}
\usepackage{graphicx}
\usepackage{textcomp}
\usepackage[table]{xcolor}
\usepackage{xcolor}
\usepackage{hyperref}

\title{\LARGE \bf
6-DoF Grasp Pose Evaluation and Optimization via Transfer Learning from NeRFs
}

\author{Gergely S\'{o}ti$^{1,2}$, Xi Huang$^2$, Christian Wurll$^1$ and Björn Hein$^{1,2}$
\thanks{The authors are with $^1$Institute of Robotics and Autonomous Systems, Karlsruhe University of Applied Sciences, 76133 Karlsruhe, Germany, and $^2$Institute for Anthropomatics and Robotics, Karlsruhe Institute of Technology, 76131 Karlsruhe, Germany.}%
\thanks{We would like to thank Prof. Andreas Wagner of the Karlsruhe University of Applied Sciences for providing computational resources for this research.}%
\thanks{This research is being conducted as part of the KI5GRob project funded by the German Federal Ministry of Education and Research (BMBF) under project number 13FH579KX9.}%
\thanks{\tt\small gergely.soti@h-ka-w.de}%
}

\begin{document}

\maketitle
\thispagestyle{empty}
\pagestyle{empty}

\begin{abstract}
We address the problem of robotic grasping of known and unknown objects using implicit behavior cloning. We train a grasp evaluation model from a small number of demonstrations that outputs higher values for grasp candidates that are more likely to succeed in grasping. This evaluation model serves as an objective function, that we maximize to identify successful grasps. Key to our approach is the utilization of learned implicit representations of visual and geometric features derived from a pre-trained NeRF. Though trained exclusively in a simulated environment with simplified objects and 4-DoF top-down grasps, our evaluation model and optimization procedure demonstrate generalization to 6-DoF grasps and novel objects both in simulation and in real-world settings, without the need for additional data. Supplementary material is available at: \url{https://gergely-soti.github.io/grasp}
\end{abstract}


\section{Introduction}
Robots are increasingly being deployed in less controlled environments, from domestic settings with vacuuming robots to medical facilities with surgical robots, and even in industrial and surveillance contexts. Despite significant research efforts in the field of robotic manipulation, robots that can handle everyday objects are still missing from real-world environments. This gap is particularly notable given the wide range of potential applications in logistics, retail, and home automation. One fundamental task that stands as a bottleneck to these applications is grasping, which is essential for activities ranging from simple pick-and-place tasks in packaging and sorting to fulfilling orders in warehouses, restocking shelves in stores, and even loading dishwashers at home.

\begin{figure}[htbp]
\centerline{\includegraphics[width=0.49\textwidth]{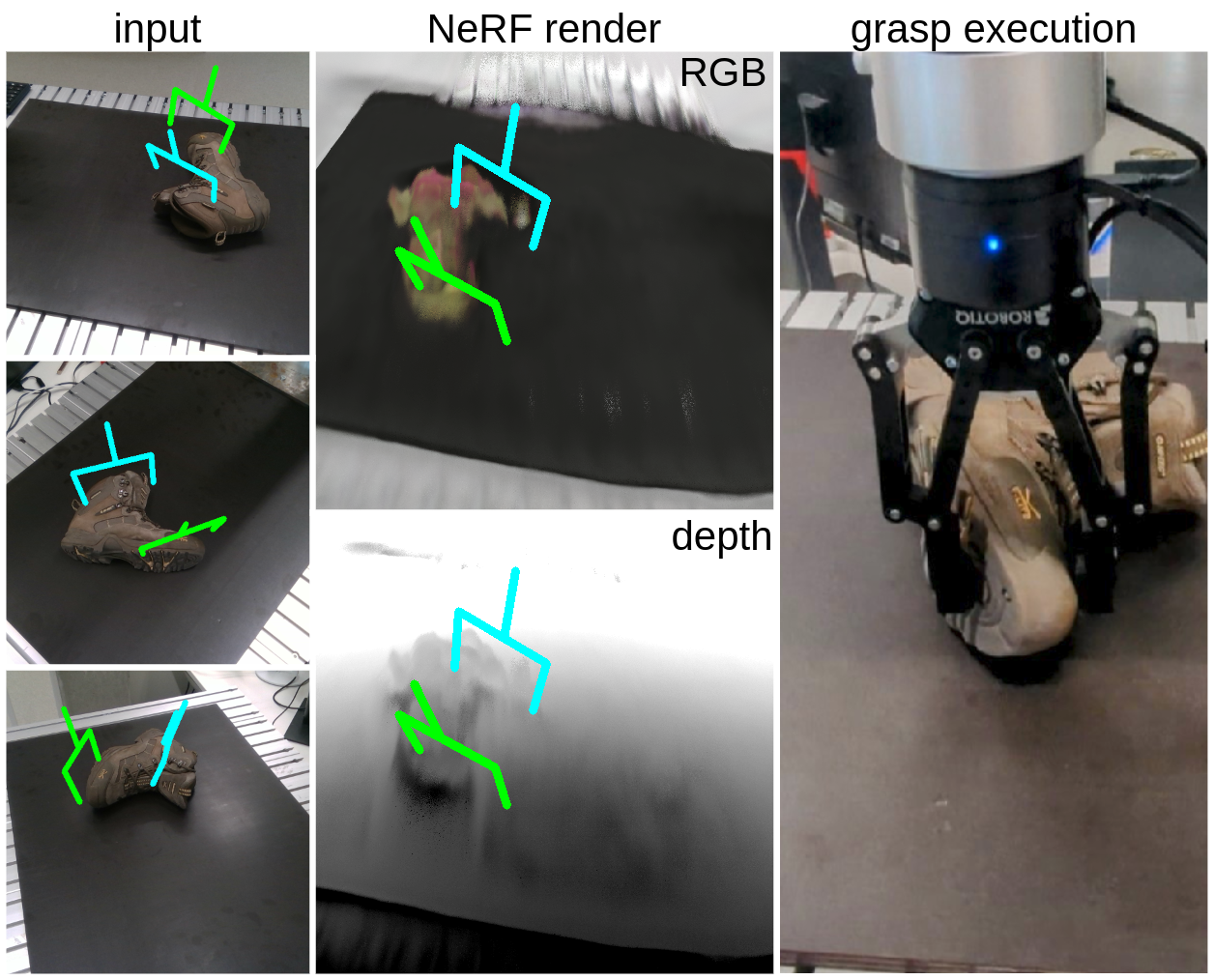}}
\caption{Two grasp candidates  after optimization for grasping a boot in the real world. The left images display the model's input, the middle images depict how the model 'imagines' the scene from a different perspective, and the right image shows the executed grasp, indicated by the green marker. Note that the markers depicting the gripper pose are overlays and do not get occluded by objects in the scene. The model was only  trained in simulation using monochromatic prismatic objects and top-down grasps; it has never been seen boots, real-world images, or non-top-down grasps.}
\label{fig:real_boot}
\end{figure}

Addressing the challenge of grasping in less controlled, real-world environments is far from trivial. Several factors contribute to this challenge: the need to generalize to unknown objects, achieving reliable performance under real-world conditions, and scaling to handle the diversity and uncertainties inherent in real-world settings. Although some existing approaches perform well in the real-world, they rely on real-world data or heavily augmented and fine-tuned simulated data \cite{levine2018learning, song2020grasping, zeng2022robotic, horvath2022sim2real, weisenbohler2022scene}. These methods would still struggle with mismatches in data distribution, a likely occurrence in real-world applications, and would require a vast amount of data to overcome this limitation.

In this work, we introduce a method that leverages Neural Radiance Fields (NeRF) for robotic grasping. We transfer the learned implicit representations of visual and geometric features from a trained NeRF model to evaluate grasp candidates, where higher values indicate a more likely successful grasp. Using the output of this function as an objective for optimization enables the identification of successful 6-degrees-of-freedom (DoF) grasp poses based on RGB camera images with known intrinsics and extrinsics. Although our model was trained on simple 4-DoF grasps in simulation, it generalizes to 6-DoF grasps both in simulation and in the real world (Fig. \ref{fig:real_boot}) without additional modules like segmentation or pose estimation. Generalization to the real-world is achieved without adaptation or using real-world training data.

\section{Related Work}

Common approaches to skill acquisition in robotics from low-level sensor data include imitation learning \cite{hussein2017imitation, fang2019survey} and reinforcement learning \cite{kober2013reinforcement, nguyen2019review}. Imitation learning leverages human demonstrations for supervised learning of a policy. Reinforcement learning operates autonomously but requires extensive data and environmental interactions, making it less practical for real-world applications. Given these constraints, imitation learning emerges as preferable when human expert demonstrations are more convenient and cost-effective.

Behavior cloning is a widely-used technique within imitation learning for robotic skill acquisition in real-world settings. Explicit models \cite{rahmatizadeh2018vision, zeng2021transporter, florence2019self, zhang2018deep} directly map observations to actions, such as converting visual cues into end-effector poses for successful grasps. Recently, implicit models have emerged as a promising alternative  \cite{florence2022implicit, soti2023gradient}. These models utilize an energy-based framework to learn a function that serves as an objective for an optimization problem, the solution to which dictates the action \cite{florence2022implicit}. Similarly, another study \cite{weng2023neural} conceptualizes grasping through an implicit function and optimization but deploys a different learning strategy, focusing on the distance to the nearest successful grasp rather than behavior cloning.

Beyond the scope of behavior cloning, implicit representations have been employed to improve or augment methods of robotic grasping \cite{ichnowski2021dex, kerr2022evo, dai2023graspnerf}. Their usefulness also extends to various other tasks such as novel view synthesis \cite{mildenhall2021nerf, yu2021pixelnerf, lin2023vision}, semantic localization \cite{kerr2023lerf}, segmentation \cite{cen2023segment}, 3D reconstruction \cite{chibane2020neural, mescheder2019occupancy}, and robot navigation \cite{adamkiewicz2022vision}.

In our method, we utilize NeRFs \cite{mildenhall2021nerf} as implicit representation to learn a scene's 3D structure using RGB images with known camera parameters. Using this representation we learn to evaluate grasp candidates, with higher values indicating a more likely successful grasp. This evaluation is used in a gradient-based optimization method similar to \cite{weng2023neural}, but we maximize the estimated value instead of minimizing the estimated distance to the nearest successful grasp pose. This requires far less data and aligns with the behavior cloning paradigm. Building on our previous work \cite{soti2023gradient}, we extend the model's applicability from 3-DoF to 6-DoF grasps. We further show that our approach is able to generalize from a small number of 4-DoF grasp demonstrations in a simulation to 6-DoF grasps in cluttered and the real-world settings.

\section{Method}
Leaning on the idea from \cite{soti2023gradient}, we use the learned geometric representation of a trained NeRF as a backbone to extract additional kinds of information from the scene. We focus on grasping, but the methodology is versatile, not being restricted to specific affordances, sensor types, or additional scene attributes. This approach builds on principles already used in novel view synthesis with the camera sensor as target.
\subsection{NeRF as a Scene Representation}
NeRFs are trained on images from different perspectives of a scene and learn a consistent geometric and visual representation of the scene \cite{mildenhall2021nerf}. They inherently learn to represent a single, static scene but there are extensions that condition NeRFs on images from different scenes and are therefore able to represent multiple scenes given some observations \cite{yu2021pixelnerf, lin2023vision}. In our approach, we use a modified version of VisionNeRF \cite{lin2023vision}: Multi-View VisionNeRF (MVNeRF)\footnote{See supplementary material for detailed architecture.\label{fn:supplementary-material}}. In contrast to VisionNeRF, it is able to process multiple input images simultaneously. MVNeRFs map the 5-DoF space of positions $x$ and view directions $q$ to $RGB$ color and  $\sigma$ \textit{density} values in a scene $S$ given observations $o$:
\begin{equation}
    \Theta(x, q, o) = (RGB, \sigma | S)
\end{equation}
An observation consists of an image $i$ and its corresponding camera parameters $M$. The function $\Theta$ emerges from a series of interconnected modules, each contributing specific computational operations to the overall model. An input image $i$ is processed by a Vision Transformer (ViT) and a fully convolutional neural network (FCN) to compute global and local image features. For each query position $x$ and view direction $q$, the corresponding image feature is computed via projection using $M$ and bilinear interpolation. The image features, $x$, and $q$ build the input of the NeRF network, consisting of 6 MLP blocks and a final fully connected readout layer, that outputs $RGB$ and $\sigma$. If multiple images are available, they are processed independently until the 3rd MLP block and are fused via average pooling. This averaged activation is then forwarded to the last 3 MLP blocks and the final readout that produces the outputs. We define \textit{scene activations} as the average pooled activation and the activations of the last MLP blocks. The model is supervised with a reconstruction loss between a ground truth image and a view rendered using conventional volume rendering as in \cite{mildenhall2021nerf, yu2021pixelnerf, lin2023vision}.
We initialize the ViT with pretrained weights from \cite{rw2019timm} and use the same training configuration as in \cite{lin2023vision}.

\subsection{Learning Grasp Evaluation}

\begin{figure*}[htbp]
\centerline{\includegraphics[width=\textwidth]{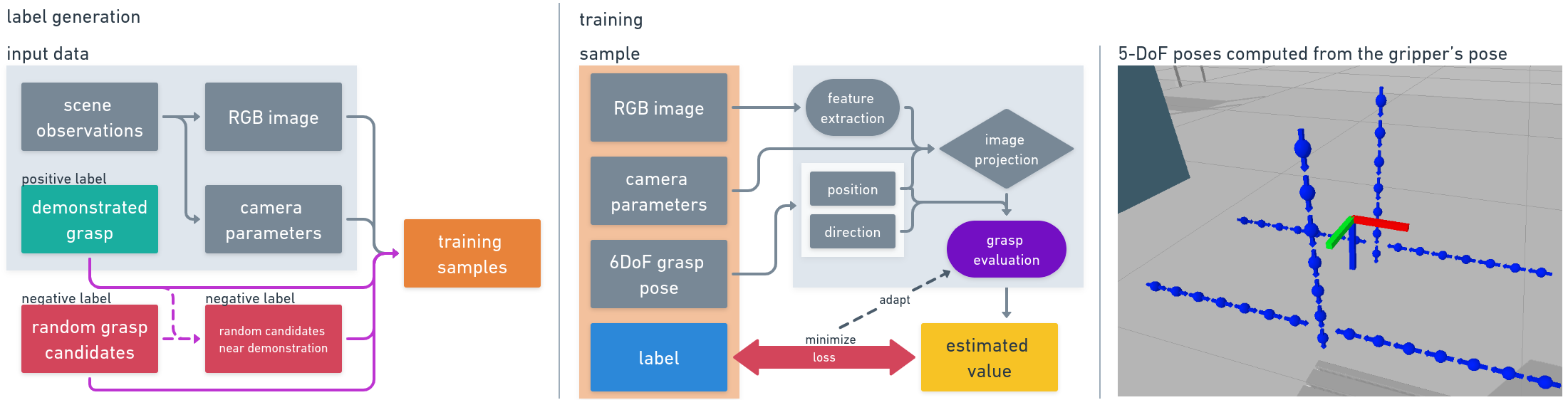}}
\caption{Left: generating training samples from a demonstration; middle and right: training method - 5-DoF poses are computed from the 6-DoF grasp candidate and are evaluated and fused in the grasp evaluation model, which incorporates the pre-trained MVNeRF.}
\label{fig:learning_grasp_eval}
\end{figure*}

We characterize a grasp by the pose of its tool center point (TCP) $p$. Given observations $o$ of scene $S$, our model should output high values in successful grasp poses and low values where nothing can be grasped, thus it should map to a value $v$ that behaves similarly to \textit{grasp probability}: the probability that 6-DoF grasp $p$ executed in $S$ results in a successful grasp. The grasp evaluation model is defined as follows:
\begin{equation}
    \Psi(p, o) = v
\end{equation}
To be able to use a trained MVNeRF, we first need to transfer the 6-DoF grasp into the domain of 5-DoF poses. For this, we compute a predefined set of position and direction pairs (5-DoF poses) relative to $p$ encompassing the TCP (Fig. \ref{fig:learning_grasp_eval}, right). Each position $x$ and direction $q$ is processed by the MVNeRF and we use the \textit{scene activations} as input to a readout neural network that fuses the information about these 5-DoF poses and ultimately computes $v$\textsuperscript{\ref{fn:supplementary-material}}. More specifically, the \textit{scene activations} of the MVNeRF are processed by a fully connected layer independently reducing their dimensionality. The outputs are concatenated and are processed by an additional fully connected layer, again to reduce the dimensionality. Finally, two MLP blocks similar to those in the NeRF network compute $v$ from these features.

We use grasp demonstrations for supervision. For each training scene, a ground truth grasp is demonstrated. This is considered as a positive sample. We then randomly sample poses in the workspace of the robot and label them as negative samples (Fig. \ref{fig:learning_grasp_eval}, left). During implementation and initial testing, we found that it is beneficial to sample some negative poses in the proximity of the ground truth grasp pose. Although there might be additional valid grasps in these samples the probability is small. We treat the problem of learning successful grasps as a classification problem and train it with cross entropy loss using these labeled samples. During training, we only update the weights of the readout neural network (Fig. \ref{fig:learning_grasp_eval}, middle) and use the Adam optimizer \cite{kingma2014adam} with a learning rate of $10^{-4}$.

\subsection{Grasp Pose Optimization}
With the learned grasp evaluation model we can define the implicit grasp model that we use to find a successful grasp pose $p^*$ in scene $S$ given observations $o$ by optimizing the objective function $\Psi$:
\begin{equation}
    p^* = \arg\max_{p} \Psi(p, o)
\end{equation}
$\Psi$ is differentiable, thus we can apply gradient based methods for optimization. To find suitable grasp poses we optimize random initial poses sampled within a predefined workspace by maximizing the output of $\Psi$ as outlined in algorithm \ref{alg:grasp-opt}.

Gradient based update of transformations can lead to invalid transformations depending on the representation and requires post-processing. During development, we explored various transformation representations. Methods describing the entire transformation, such as transformation matrices, screw notation, and forward kinematics were less successful than approaches which decouple position and orientation. Though orientation remains challenging to represent, rotation vectors, quaternions and the axis-angle representation performed similarly. However, we opted for quaternions because they showed more consistent performance. Quaternions, being unit vectors, are normalized during post-processing. Additionally, position values are clipped if they venture outside the predefined workspace due to gradient ascent.

\begin{algorithm}
\caption{
\small Grasp pose optimization}
\begin{algorithmic}[1]
\small
\Require Observation $o$
\Ensure Successful grasp $p^*$
\State $G \gets \text{RandomGraspCandidates()}$ \Comment{Initialization}
\While{Not Terminate} \Comment{Termination criterion}
    \ForAll{$p \in G$} \Comment{Parallelized}
        \State $p \gets p + \nabla \Psi(p, o)$  \Comment{Maximize $\Psi$}
        \State $p \gets \text{PostProcess}(p)$  \Comment{Fix pose}
    \EndFor
\EndWhile
\State $p^* \gets \arg\max_{p \in G} \Psi(p, o)$  \Comment{Grasp with highest success}
\State \Return $p^*$
\end{algorithmic}
\label{alg:grasp-opt}
\end{algorithm}

In the gradient-based update step, we use the Adam optimizer \cite{kingma2014adam} to separately optimize position and orientation. The optimization process uses a fixed number of iterations as its termination criterion. Initially, 16 optimization steps are executed to adapt only the positions, utilizing an initial learning rate of 0.05 and an exponential decay rate of 0.9. Subsequently, we conduct another set of 16 optimization steps to optimize only orientation, with an initial learning rate of 0.05 and an exponential decay rate of 0.99. While this is our current approach, it is conceivable to use more sophisticated methods that optimize both position and orientation simultaneously or dynamically alter the optimization strategy based on plateaus in the objective function or specific convergence thresholds.

\section{Results}
We evaluated our proposed robotic grasping model on three simulated tasks and one real-world task. The simulated tasks were designed to assess the model's performance under varying levels of complexity and uncertainty:
\begin{itemize}
    \item \textbf{simple:} the workspace contains up to five monochromatic prismatic objects, each placed at a considerable distance from one another. The task is to grasp one of the objects.
    \item \textbf{clutter:} the workspace contains five monochromatic prismatic objects randomly placed in a cluttered arrangement. The task is to grasp all objects successively. 
    \item \textbf{novel objects:} the workspace contains one unseen object selected from the YCB dataset \cite{calli2015ycb}. The task is to grasp this novel object.
\end{itemize}
For the \textbf{real-world} task, various everyday objects were deliberately positioned in a physical workspace to validate the model's real-world applicability. A grasp was deemed successful across all tasks if the gripper enveloped the object and lifted its center of mass. Illustrative examples from these tasks are depicted in Fig. \ref{fig:real_boot} and \ref{fig:sim_renders}.

For training, all models were exclusively exposed to the \textbf{simple} scenario. Importantly, the training dataset does not feature objects in close proximity, with complex textures or in orientations divergent from an upright position. NeRFs are trained with 2.5k scenes with images from 50 random perspectives for 500k epochs, grasp success estimators are trained with 512 scenes each with a single demonstrated grasp for 400 epochs. The training of NeRFs took around 5 days and the training of grasp success estimators a couple of hours on an NVIDIA RTX A6000.

We use the attention module of Transporter Networks (TN), as defined in \cite{soti2023train} as a baseline. This choice was motivated by its proven effectiveness in pick-and-place tasks using limited training data similar to our training setup, although it also requires depth information. The original Transporter Network architecture \cite{zeng2021transporter} proposes an approach for inferring 6-DoF gripper poses for object placement. Adapting this approach to grasping consists of estimating an SE(2) grasp pose using the attention module of \cite{soti2023train} and regressing the remaining $z$-coordinate and roll and pitch angles. Regression of additional rotational parameters was deemed infeasible due to the orientation limitations of our training data. Regarding the $z$-coordinate, we found that estimation methods were inaccurate, prompting us to rely on calculating the $z$-coordinate from the depth value at the SE(2) grasp location derived from the input RGB-D images. Additionally to the Transporter network we also train grasp success estimators of the same architecture as $\Psi$ but without pre-trained MVNeRFs. All baselines are trained for 400 epochs and took a couple of hours to complete.

\subsection{MVNeRF}

\begin{figure}[htbp]
\centerline{\includegraphics[width=0.45\textwidth]{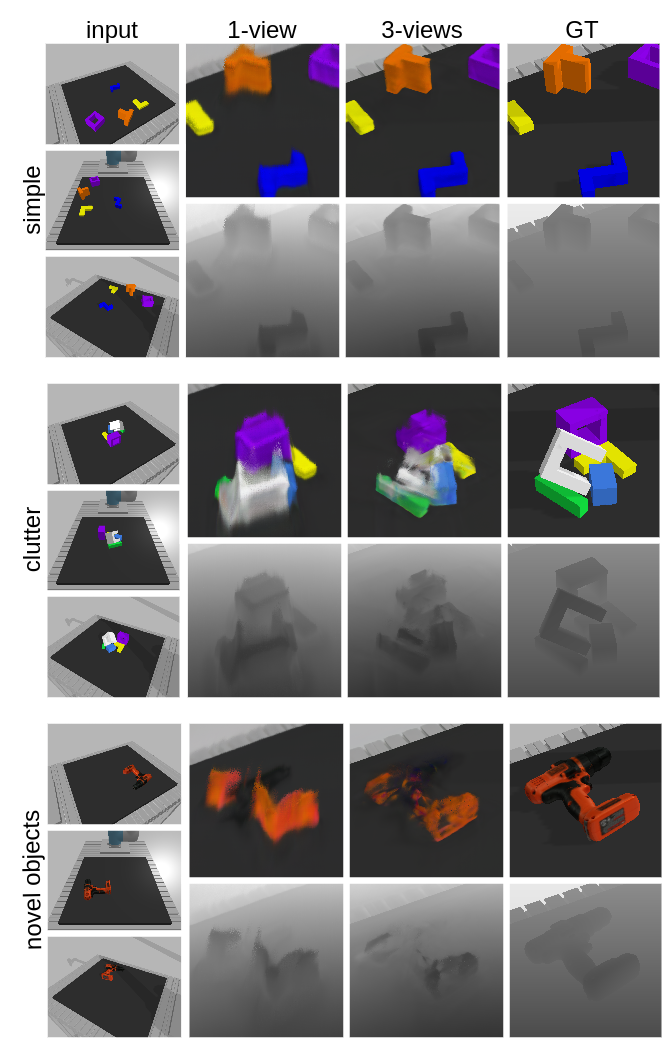}}
\caption{MVNeRF renderings for the simulated tasks. Left: input images with known camera parameters; middle: rendering of 1-view and 3-views MVNeRF; right: ground truth generated in simulation. Note that depth information was neither utilized in training nor in inference, and is displayed only for visualization.}
\label{fig:sim_renders}
\end{figure}

We trained two variants of MVNeRF: a single-view version (1-view), functionally equivalent to VisionNeRF \cite{lin2023vision}, and a multi-view version (3-views) that processes three images concurrently. Both variants were applied in simulated and real-world tasks (see Fig. \ref{fig:sim_renders} and Fig. \ref{fig:real_boot}). In the \textbf{simple} task, both models produced highly recognizable renderings. However, the 3-views model demonstrated superior levels of quality and detail in both RGB and depth images. It should be noted that while depth images can be rendered using NeRFs, they were neither used during the training nor as part of the input. In the \textbf{clutter} task, both models retained the general shape of the clutter, but the colors were blended, making the objects indistinguishable. The most significant difference between the two models appeared in the \textbf{novel objects} task: while the 1-view model generated blurry approximations of the objects, the 3-views model rather captured their surface details. In the \textbf{real-world} task, both models produced blurry but recognizable color images. But the depth image rendered by 1-view captured the objects significantly better than the depth from 3-views. These effects are most likely caused by sensor noise and inaccuracies in the camera's calibration. In this work, we primarily focus on using NeRFs for grasping; a comprehensive quantitative analysis of MVNeRFs is beyond the current scope.

\subsection{Grasp Evaluation}

\begin{figure*}[htbp]
\centerline{\includegraphics[width=\textwidth]{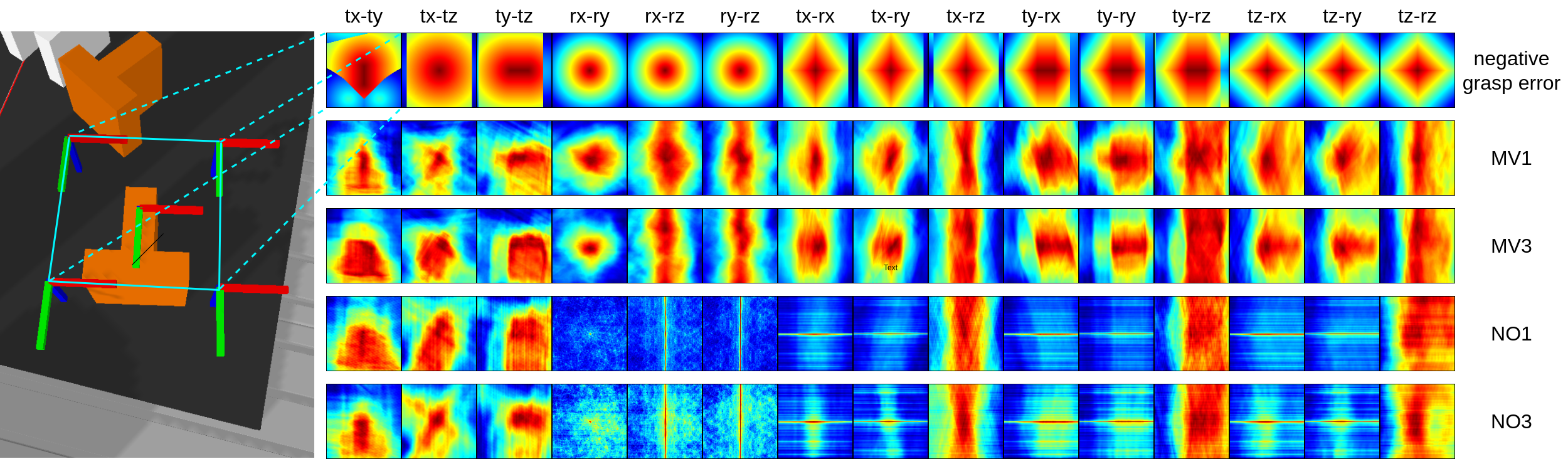}}
\caption{Depiction of grasp value estimations in 6-DoF space. Left: a valid grasp pose and the corners of the tx-ty slice (translational displacement along x and y axes); right: visualization of negative grasp errors and estimated grasp values across multiple slices in translational and rotational dimensions. Interpretation of the slices: in the tx-ty slice of the negated grasp error function, regions maintaining red hues signify deviations along the y-axis still result in valid grasps. Bright white spots, corresponding to the arms of the T-shaped object, are indicative of minimal translational error, yet are not red due to maximal rotational error.
}
\label{fig:grasp_value}
\end{figure*}

We trained four 6-DoF grasp evaluation models employing the $\Psi$ architecture for implicit grasp pose estimation. Two models utilized one view (MV1) and three views (MV3) with their respective pre-trained MVNeRFs (1-views and 3-views). Additionally, we trained two models without pre-trained NeRFs, one using a single view (NO1) and another using three views (NO3). In case of MV1 and NO1, we processed all three views available to MV3 and NO3 independently and averaged their results to obtain the final value as suggested in in \cite{soti2023gradient}.

To ascertain the validity of our grasp evaluation, we define a surrogate function for \textit{grasp probability}: the \textit{negative grasp error}. This function computes translation and rotation errors relative to the nearest valid grasp pose, averages them, and negates the result. The nearest grasp pose is determined via an oracle with knowledge of object poses and valid grasps within the simulation as described in \cite{soti2023gradient}. Although this is an approximation, it serves as a useful heuristic for showing the characteristics of the real \textit{grasp probability}.

Figure \ref{fig:grasp_value} illustrates slices of both the \textit{negative grasp errors} and the outputs of our trained models centered on a successful grasp pose. Our findings indicate that models equipped with pre-trained MVNeRFs yield estimates more similar to the \textit{negative grasp errors}, particularly in rotational dimensions. This observation is exemplified in a T-shaped object, where movement along the y-axis preserves grasp validity, as illustrated in the tx-ty, ty-tz, ty-rx, ty-ry, and ty-rz slices. 

Our results suggest an implicit approximation of the set of successful grasps is attainable. Additionally, the gradient properties of the slices indicate suitability for gradient-based optimization methods, particularly for MV1 and MV3, whereas NO1 and NO3 displayed satisfactory properties only in translation slices.

\subsection{Grasping in Simulation}
The highest grasp success rate in the \textbf{simple} task was recorded for MV1 (0.88), followed closely by MV3 and TN (see Table \ref{table:sim_results}). Both NO1 and NO3 displayed notably inferior performance. In this task we had access to the oracle, thus we also computed the translation and rotation errors. Despite TN's significantly lower translation and rotation errors (averages of 0.8mm and 3.2°), its performance lagged slightly behind MV1 (with 4.3mm and 10.7° errors on average). This discrepancy could likely be attributed to two key factors: (1) the objects in this task were geometrically simple, and thus more forgiving of grasping errors; (2) the error function is not a direct surrogate for grasp success, meaning that minimal errors at specific poses may still lead to grasp failures, e.g. at the junction of the arms and the stem of the T-shape, the gripper could collide with the object during approach and fail to grasp it.

{\rowcolors{7}{gray!10}{gray!10}
\begin{table}
\begin{center}
\caption{\label{table:sim_results} Results in simulated tasks: grasp success rates for all tasks, average number of objects cleared, and average number of objects dropped in the clutter task}
\begin{tabular}{l || c|| c | c| c || c }
\multicolumn{1}{c ||}{}& \textbf{simple} & \multicolumn{3}{c ||}{\textbf{clutter}} & \textbf{novel objects} \\
\cline{2-6}
\multicolumn{1}{c ||}{}& success & success & cleared & dropped & success \\
\hline
TN & 0.84 & \textbf{0.58} & 4.2 & \textbf{0.2} & 0.35  \\
NO1 & 0.4 & 0.24 & 2.4 & 1.2 & 0.07   \\
NO3 & 0.4 & 0.12 & 1.2 & 1.2 & 0.11 \\
MV1 & \textbf{0.88} & \textbf{0.58} & \textbf{4.4} & 0.4 & 0.44  \\
MV3 & 0.84 & 0.57 & 4.0 & 1.0 & \textbf{0.50} \\
\end{tabular}
\end{center}
\end{table}

In the \textbf{clutter} task, each model had 10 attempts to clear a cluttered scene consisting of 5 objects. Their performance was measured by their grasp success rate, the number of objects successfully cleared from the scene, and the number of objects inadvertently removed from the workspace (e.g., dropped from the table). The results are shown in Table \ref{table:sim_results}. MV1 and TN achieved the highest success rates; however, MV1 demonstrated a somewhat better capability in clearing objects, while TN dropped slightly fewer objects. The performance of MV3 lagged behind that of MV1 and TN, and both NO1 and NO3 demonstrated significantly poorer results.

The strong performance of TN, a model restricted to 4-DoF grasps, suggests that the \textbf{clutter} task was relatively straightforward using top-down grasping strategies. This is likely attributable to the geometric simplicity of the objects, as well as their potential to land in favorable orientations following either failed or successful grasp attempts. Across all models, the majority of failed attempts occurred in proximity to other objects, subsequently altering the scene's configuration.

The \textbf{novel objects} task presented a higher level of difficulty. The task involved grasping one of the following YCB objects: banana, foam brick, gelatin box, hammer, Master Chef can, pear, power drill, strawberry and tennis ball. These objects did not only differ substantially in shape from the prismatic objects in the training data but also feature complex textures as opposed to the monochromatic objects used for training. Despite these complexities, models MV1 and MV3 were notably effective, achieving grasp success rates of 44\% and 50\%, respectively as shown in Table \ref{table:sim_results}. TN, restricted to 4-DoF grasps, demonstrated lower effectiveness and models NO1 and NO3 managed to grasp only a small fraction of the novel objects. 

Objects with simple geometries and colors like the banana and the hammer were reliably grasped by MV1, MV3 and TN. Surprisingly TN failed to grasp the tennis ball every time, while both MV models were very good at it. All models failed at grasping objects with more complex textures, like the Master Chef can and the gelatin box. The foam brick was also only grasped rarely, likely due to its dark color blending into the ground. The primary reason for unsuccessful grasps for both MV1 and MV3 was object slippage within the gripper whereas the failures of TNs were primarily due to collisions.

The results indicate that models MV1 and MV3 capture a more generalized representation of robotic grasping, and show the utility of leveraging pre-trained Neural Radiance Fields (NeRFs) as a foundational architecture for implicit grasp models. The fact that MV1 mostly outperforms MV3 warrants further investigation; possible explanations could include the single-image-trained NeRF's necessity for stronger object shape generalization, or an ensemble effect produced by combining the outputs for three different inputs during grasp evaluation.

\subsection{Real-World Grasping}

The real-world setup, consisting of a workbench and a UR10e robot, is similar to the simulation but has two key differences: it employs a RobotiQ 2F-140 gripper rather than the simulated 2F-85, and it includes an Intel RealSense D415 camera on the robot's wrist for data collection.

For the \textbf{real-world} task, we selected a variety of everyday objects: tennis ball, crochet ball, small Lego tire, large Lego tire, can of beans, rubber duck, hiking boot, power drill, shampoo bottle and a 3d printed block. Each object was subjected to four trials. During each trial, the object was randomly placed within one of four designated quadrants on the workbench. We only tested models trained in simulation. Models NO1 and NO3 were excluded from real-world testing due to poor performance in simulated trials.

Among the tested models, only MV1 was able to successfully execute grasps. However, it failed to grasp dark-colored Lego tires that blended with the environment, consistent with the findings of the \textbf{novel objects}. Aside from these, MV1 achieved a success rate exceeding 50\%. The tennis and crochet balls were consistently grasped in all trials. The can of beans, rubber duck, hiking boot, and 3D-printed block were grasped in half of the trials, while the shampoo bottle and power drill were successfully grasped only once of four times. Failures often occurred in close spatial proximity to the target object. The most common failure mode was slight misplacement of the grasp, leading to slippage, particularly with heavier objects like the drill, boot, shampoo bottle, and can of beans.

Although some grasp attempts by the MV3 model seemed promising, they ultimately failed due to too low z-axis positioning, causing collision between the gripper and the workbench. Other attempts strayed off to the workspace boundaries. We hypothesize that these inconsistencies may arise from camera calibration inaccuracies, as the MVNeRFs model relies heavily on both extrinsic and intrinsic camera parameters for representing visual and geometric features.

TNs consistently initiated grasp attempts in a single corner of the workspace. This behavior may stem from the model's strong reliance on visual features and a likely substantial mismatch in the data distribution between real-world and simulated images.

The MV1 model's successful real-world grasps demonstrate the effectiveness of our implicit approach, which relies on NeRFs for geometric and visual feature representation. Yet, MV3's consistent failures suggest the method's sensitivity to camera calibration. Both MV1 and MV3 process three images but differently: MV1 treats each image independently, like an ensemble model, while MV3 combines information from all three. This difference could explain the varying impact of camera calibration errors on their performance.

\section{Conclusion}

In this work, we demonstrated that implicit model architectures, supported by a pretrained NeRF backbone, are effective in learning robotic grasping tasks. Our approach displayed robust generalization capabilities, scaling from 4-DoF to 6-DoF grasps across different objects and tasks. Furthermore, it demonstrated transferability from simulation to real-world conditions without any prior exposure to such data.

While our method yields promising results, it is not without constraints. The need to pre-train the NeRF backbone imposes an initial computational burden, and our model's efficacy is intrinsically tied to NeRF's capabilities in scene representation as well as the quality of the grasp optimization landscape. Furthermore, the confined workspace in our experiments raises questions not just about generalization, but also about the limitations of NeRF's capability to represent larger or more complex scenes. Our experimental setup is also geared towards grasping in general, lacking task-specific goal orientation, thereby narrowing the scope of its applicability.

Despite these limitations, our model's ability to generalize from 4-DoF to 6-DoF grasps and from simulation to real-world settings implies that it could learn more complex tasks, possibly requiring only a modest dataset for training. Future work could involve integrating task-specific goals, language-conditioning, or a closed-loop control system for more precise manipulation. Additionally, the geometric features provided by NeRFs can be leveraged for motion planning, offering an additional way to improve robotic capabilities.

Lastly, one of the unique aspects of our approach is its potential interpretability. Should the model encounter failure cases, the internal rendering capabilities of NeRFs could provide valuable insights into the underlying causes.

\bibliographystyle{IEEEtran}
\bibliography{main}

\end{document}